\documentclass[conference]{IEEEtran}

\usepackage{hyperref}
\usepackage{cite}
\usepackage{amsmath,amssymb,amsfonts}
\usepackage{cases}
\usepackage{algorithmic}
\usepackage{graphicx}
\usepackage{textcomp}
\usepackage{xcolor}
\usepackage{booktabs}
\usepackage{colortbl}
\usepackage{balance}

\usepackage[linesnumbered,ruled,vlined]{algorithm2e}
\SetKw{Continue}{continue}
\SetKw{Or}{or}
\usepackage{enumitem}
\usepackage{tikz}

\DontPrintSemicolon
\usepackage{multirow}
\usepackage{nicefrac}
\usepackage{tabularx}
\usepackage{mathtools}
\usepackage{pifont}

\usepackage{siunitx}
\usepackage{subcaption}
\usepackage{microtype}
\usepackage{makecell}
\usepackage{listings}
\usepackage{color}

\newcommand*{\belowrulesepcolor}[1]{%
	\noalign{%
		\kern-\belowrulesep
		\begingroup
		\color{#1}%
		\hrule height\belowrulesep 
		\endgroup
	}%
}
\newcommand*{\aboverulesepcolor}[1]{%
	\noalign{%
		\begingroup
		\color{#1}%
		\hrule height\aboverulesep
		\endgroup
		\kern-\aboverulesep
	}%
}

\lstdefinestyle{customc}{
  belowcaptionskip=1\baselineskip,
  breaklines=true,
  frame=L,
  xleftmargin=\parindent,
  language=C,
  showstringspaces=false,
  basicstyle=\footnotesize\ttfamily,
  keywordstyle=\bfseries\color{green!40!black},
  commentstyle=\itshape\color{purple!40!black},
  identifierstyle=\color{blue},
  stringstyle=\color{orange},
}

\lstdefinestyle{customasm}{
  belowcaptionskip=1\baselineskip,
  frame=L,
  xleftmargin=\parindent,
  language=[x86masm]Assembler,
  basicstyle=\footnotesize\ttfamily,
  commentstyle=\itshape\color{purple!40!black},
}

\ifCLASSINFOpdf
\else
\fi

\begin{document}
\title{ReconROS Executor: Event-Driven Programming of FPGA-accelerated ROS 2 Applications}

\author{
	\IEEEauthorblockN{Christian Lienen}
	\IEEEauthorblockA{\textit{Paderborn University}\\
	\textit{Germany}\\
	christian.lienen@upb.de}
\and
	\IEEEauthorblockN{Marco Platzner}
	\IEEEauthorblockA{\textit{Paderborn University}\\
	\textit{Germany}\\
	platzner@upb.de}
}

\maketitle

\begin{abstract}
Many applications from the robotics domain can benefit from FPGA acceleration. A corresponding key question is how to integrate hardware accelerators into software-centric robotics programming environments. Recently, several approaches have demonstrated hardware acceleration for the robot operating system (ROS), the dominant programming environment in robotics. ROS is a middleware layer that features the composition of complex robotics applications as a set of nodes that communicate via mechanisms such as publish/subscribe, and distributes them over several compute platforms.

In this paper, we present a novel approach for event-based programming of robotics applications that leverages ReconROS, a framework for flexibly mapping ROS 2 nodes to either software or reconfigurable hardware. The ReconROS executor schedules callbacks of ROS 2 nodes and utilizes a reconfigurable slot model and partial runtime reconfiguration to load hardware-based callbacks on demand. We describe the ReconROS executor approach, give design examples, and experimentally evaluate its functionality with examples.
\end{abstract}

\IEEEpeerreviewmaketitle

\section{Introduction}
\label{sec:Introduction}

Many robotics applications are computationally very demanding, in particular when they process large amounts of data sensed from their environment and run involved algorithms to compute state information and next actions to take. In the last years, efficient implementations of robotics applications on high-performance embedded platforms comprising multi-core CPUs, general-purpose GPUs, or FPGAs have been studied and several works have shown the potential advantage of FPGAs over CPUs and GPUs with respect to performance, energy consumption, and latency, e.g.,\cite{8782524, 7298356, 7577310, 8401525}. 

A key question for FPGA acceleration of robotics applications is how to integrate hardware accelerators into software-centric robotics programming environments. Recently, several approaches targeted the robot operating system (ROS), which is the dominant programming environment in robotics. ROS is essentially a middleware layer that allows for the decomposition of complex robotics applications into a set of nodes that communicate via mechanisms such as publish/subscribe. The nodes can then be distributed over several compute platforms. Most approaches partition a ROS node and map the computation intensive kernels to reconfigurable hardware. A few approaches go further and support the mapping of complete ROS nodes to hardware, which greatly increases flexibility and facilitates design space exploration. However, the mapping of ROS nodes to either software or hardware is static and event-driven programming, which is central to ROS 2, the latest version of ROS, is not supported.

In this paper, we present the ReconROS executor for event-driven programming of ROS 2 applications with flexible hardware acceleration. The ReconROS executor registers ROS 2 node functions as callbacks and dispatches them to run in software on one of several processor cores or in hardware in the reconfigurable fabric. While the executor builds on the architecture and functionality of ReconROS~\cite{LP20,lienen2021design}, a previously presented open source framework for hardware acceleration in robotics, the main novelty is that the ReconROS executor employs partial hardware reconfiguration with a reconfigurable slot model to load and execute hardware-mapped ROS 2 callbacks on demand, and that it can dispatch callbacks for either software or hardware execution. As a result, ROS 2 developers can exploit the benefits of hardware acceleration from their standard programming environment.

The remainder of the paper is organized as follows: Section~\ref{sec:BackgroundRelatedWork} provides background, in particular an overview over ROS and the ROS executor for event-driven programming, and a discussion of related work that aims at integrating FPGA hardware acceleration with ROS. Section~\ref{sec:ReconROSExecutor} explains the hardware architecture and design concept for the novel ReconROS executor, before Section~\ref{sec:designexample} presents an example for the configuration and coding steps when developing a ReconROS application. Section~\ref{sec:Exp} reports on experiments to demonstrate the functionality and advantages of our approach. Finally, Section~\ref{sec:ConclusionFutureWork} concludes the paper and gives an outlook to future work.

\section{Background and Related Work}
\label{sec:BackgroundRelatedWork}

In this section, we first introduce to the robot operating system ROS, then focus on the functionality of the ROS executor and, finally, review related approaches for making hardware acceleration available to ROS-based robotics applications.

\subsection{The Robot Operating System}
\label{sec:BackgroundRelatedWork:ros2}

The Robot Operating System (ROS) is the dominant framework for robotics programming. ROS comprises a middleware layer for data communication within a computer network and tools and libraries for rapid and modular development of large and complex applications. ROS decomposes an application into a set of ROS nodes that can communicate by exchanging messages using three possible mechanisms: An $m$:$n$ publish-subscribe mechanism and two $1$:$1$ mechanisms denoted as services and actions, that follow a client-server model. Services allow for nodes to access functionalities of other nodes in a manner similar to remote procedure calls, and actions are more elaborate and combine two service requests with a publish-subscribe communication for regular feedback information. 

\begin{figure}[!h]
	\center
    \includegraphics[width=1\linewidth]{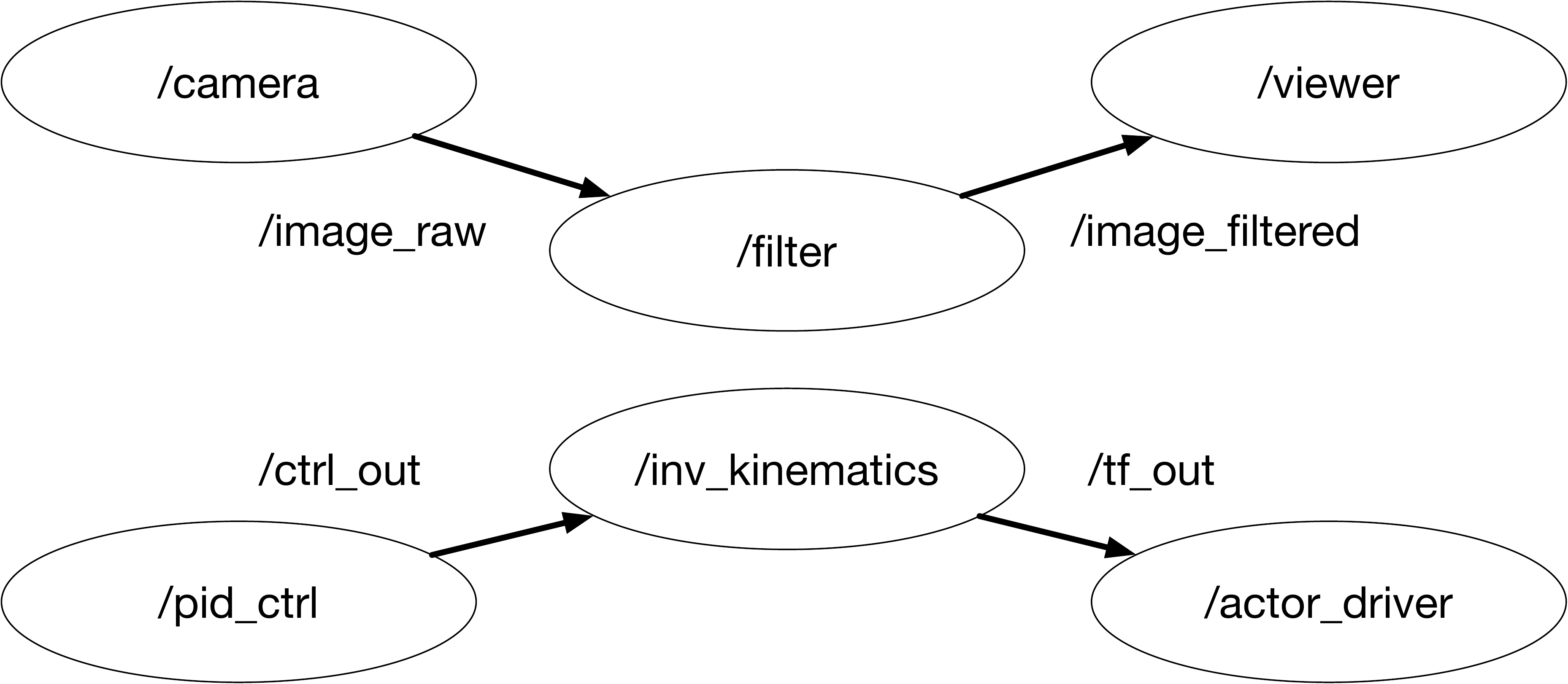}    
    \caption{ROS computational graph example}
    \label{fig:computational_graph}
\end{figure}

ROS applications are often represented as computation graphs, in which the nodes represent ROS nodes and the edges represent a ROS-supported form of communication~\cite{ros_computational_graph}. The nodes of a computation graph are then mapped to compute platforms in a distributed system. Figure \ref{fig:computational_graph} outlines an example for a computation graph using publish-subscribe communication. The ROS node {\tt /camera} captures images from a camera and publishes them to the topic {\tt /image\_raw}. The sobel {\tt /filter} node subscribes to this topic and publishes the filtered images to the topic {\tt /image\_filtered}. The {\tt /viewer} node displays the filtered images. The second part of the application implements a control loop for servo control, where the node {\tt /pid\_ctrl} runs a PID control algorithm on position sensor input, the node {\tt /inv\_kinematics} determines the required new position and, finally, the node {\tt /actor\_driver} sets the motor signals accordingly.

The most recent version ROS 2 builds on several layers on top of the operating system Linux. As communication layer, ROS 2 uses the data distribution service (DDS), a common standard for publish-subscribe communication specified by the Object Management Group (OMG)~\cite{dds_omg}. The ROS middleware layer (rmw) in combination with a DDS implementation-specific adaptor in the ROS 2 stack allow for interchangeability of different DDS implementations. There are several different DDS implementations available, e.g., the eProsima FastRTPS or RTI Connext DDS. Additionally, shared-memory based communication between nodes is enabled by, for example, iceoryx. On a higher level, the rcl library comprises the standard framework for providing ROS concepts. High-level libraries (rclcpp for C++ or rclpy for Python) wrap rcl and provide more advanced functionalities, e.g., execution management. 

Another key part of the ROS framework is the message infrastructure. Messages can be hierarchically composed out of basic built-in elements such as integers or floats using a common description language. Further, ROS includes standard message packages, e.g., for images or positional data. The ROS tool flow comprises tools for translating such message definitions into data representations suitable for inclusion in high-level language programs in, for example, Python and C++.

\subsection{The ROS 2 Executor}
\label{subsec:ROS2Executor}

The ROS nodes mapped to one computation platform can execute as Linux processes or threads using the underlying Linux scheduler. In such a case the nodes need to regularly poll the communication layer for available messages. However, the more common model under ROS 2 is the event-driven model, where nodes register callbacks that are executed when specific events occur. There are four categories of callbacks: Callbacks executed by any node when a (periodic) timer event occurs, callbacks executed by a subscriber on a received message, callbacks executed by a ROS 2 service server on a received service request, and callbacks executed by a ROS 2 service client on a received service response. 

ROS 2 provides a so-called executor function that interacts with the underlying communication layer and timer infrastructure to catch events and execute callbacks in a run-to-completion mode utilizing one or more worker threads. That is, callbacks are not preempted.

\begin{figure}[ht]
	\center
    \includegraphics[width=0.95\linewidth]{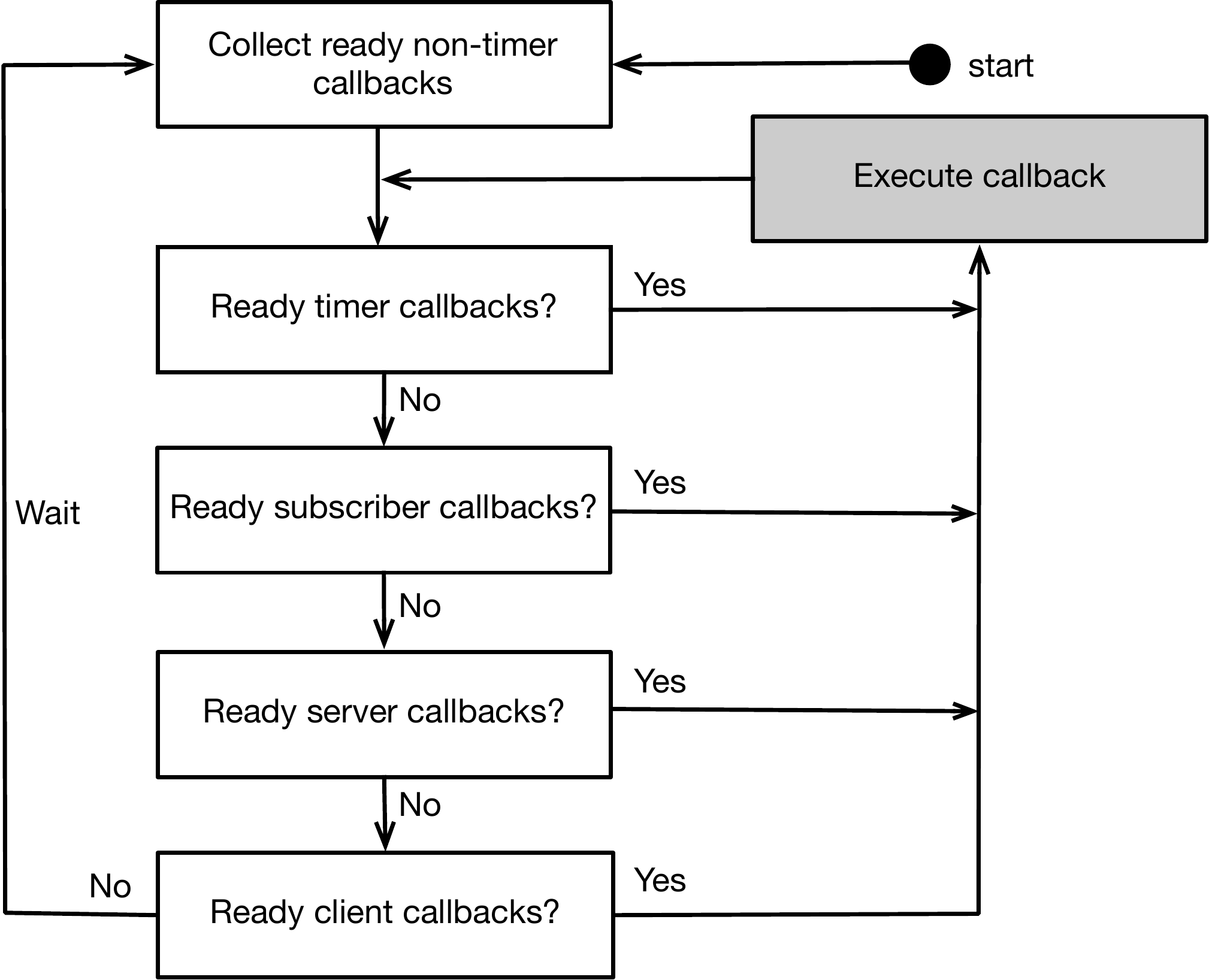}    
    \caption{ROS 2 standard executor scheduling algorithm~\cite{casini2019response}}
    \label{fig:ros2_standard_executor}
\end{figure}

By default, ROS 2 offers standard single-threaded and multi-threaded executors for C++ and Python applications that implements the scheduling algorithm sketched in Figure~\ref{fig:ros2_standard_executor}. The algorithm comprises two nested loops. In the outer loop, the executor interacts with the DDS layer to collect all ready subscriber, server, and client callbacks into a {\it readySet}. In the inner loop, the executor checks for timer-triggered callbacks and, if such are available, executes them. Then, subscriber, server, and client callbacks are considered in that order and if such a callback is ready, it is executed and removed from the {\it readySet}. If there a no more callbacks ready, the next iteration of the outer loop is started after a configurable waiting time. 

The ROS 2 executor implicitly implements priorities in the sense that timer-triggered callbacks get high priority, and the other callbacks lower priorities, since they are first collected in the outer loop and then executed in the inner loop in the order shown in Figure~\ref{fig:ros2_standard_executor}. Within one callback category, requests are ordered by the sequence of their initial registration at the executor.

The real-time behavior of the ROS 2 executor was studied in~\cite{casini2019response}. The authors analyzed the response time of  ROS 2 applications and provided a scheduling model, a worst-case response time analysis, and general insights into the real-time behavior of ROS 2.
In follow-up work~\cite{9355523} the precision of the response time analysis was improved.  
There are also alternatives to the standard ROS 2 executor. For example, in~\cite{rclcexecutor} an executor for micro ROS platforms~\cite{rclcexecutor} equipped with embedded micro controllers was presented. This executor is fully coded in C, supports domain-specific requirements, and improves on the analysis of real-time aspects.

\subsection{Related Approaches for ROS-FPGA Integration}
\label{sec:BackgroundRelatedWork:relatedwork}

In the last years, several approaches have been presented for integrating reconfigurable hardware accelerators into the ROS architecture. Most of these approaches partition a ROS software node and map the runtime-intensive parts as acceleration kernels to reconfigurable hardware, while the remaining parts stay on the CPU. The works~\cite{Li2015, Yamashina2016, Sugata2017, 8823798} follow this model and focus on the automated generation of the interfaces between software and acceleration kernels, the minimization of communication time between them, and the use of high-level synthesis to increase productivity. The ReconfROS framework~\cite{reconfros} implements shared-memory communication between the software and the acceleration kernels to further reduce communication effort.

A different approach is followed in~\cite{Podlubne2020, Podlubne2021}, where one or more complete ROS nodes can be mapped to hardware. All these hardware-mapped ROS nodes connect to an AXI-based gateway followed by a protocol generator and a TCP/IP interface to allow for communication with other ROS nodes of the application. A central manager coordinates the communication between the hardware-mapped ROS nodes and the gateway. 

ReconROS~\cite{LP20,lienen2021design} also allows for mapping complete ROS nodes to hardware and combines the reconfigurable hardware operating system ReconOS~\cite{Luebbers_Platzner_2009, 6636314} with ROS 2. ReconOS features multithreaded programming with hardware and software threads, i.e., both hardware and software threads use operating system services, such as semaphores and mutexes, in exactly the same way. Additionally, all threads can access the common virtual shared memory address space. ReconROS extends the functionality of ReconOS by adding ROS 2 primitives, turning complete ROS 2 nodes into hardware threads, and allowing them to use the ROS communication mechanisms publish-subscribe, services, and actions. Since the ReconROS architecture and build system are available in open source, we have been using it as starting point for the work presented in this paper.

Industry has also taken up the ROS-FPGA integration. For example, Xilinx develops the KRIA robotics stack\cite{mayoral2021adaptive, mayoral2021kria}, which merges ROS build tool flows with the Vitis software platform. There, compute intensive calculations are outsourced to reconfigurable logic as so-called acceleration kernels.

\section{{\sc The ReconROS Executor}}
\label{sec:ReconROSExecutor}

In the existing ReconROS framework, ROS 2 hardware nodes have to be statically placed in reconfigurable logic where they remain until the application terminates. The hardware nodes run in $while(1)$-loops, that start with blocking reads for new input data, process the data, and write the output. As main novelty we introduce partial hardware reconfiguration with a reconfigurable slot model to be able to load and execute hardware-mapped ROS 2 nodes on demand, and we devise a ROS 2 executor that can dispatch callbacks for either software or hardware execution, if hardware versions of the callbacks are available. 

As a result, (i) robotics application developers can exploit hardware acceleration from their known programming environment and event-driven programming model, and (ii) the limited hardware resources are operated in an efficient manner. In the following, we discuss the hardware architecture and the design concept of the ReconROS executor.

\subsection{Hardware Architecture}

\begin{figure}[!h]
	\center
    \includegraphics[width=1\linewidth]{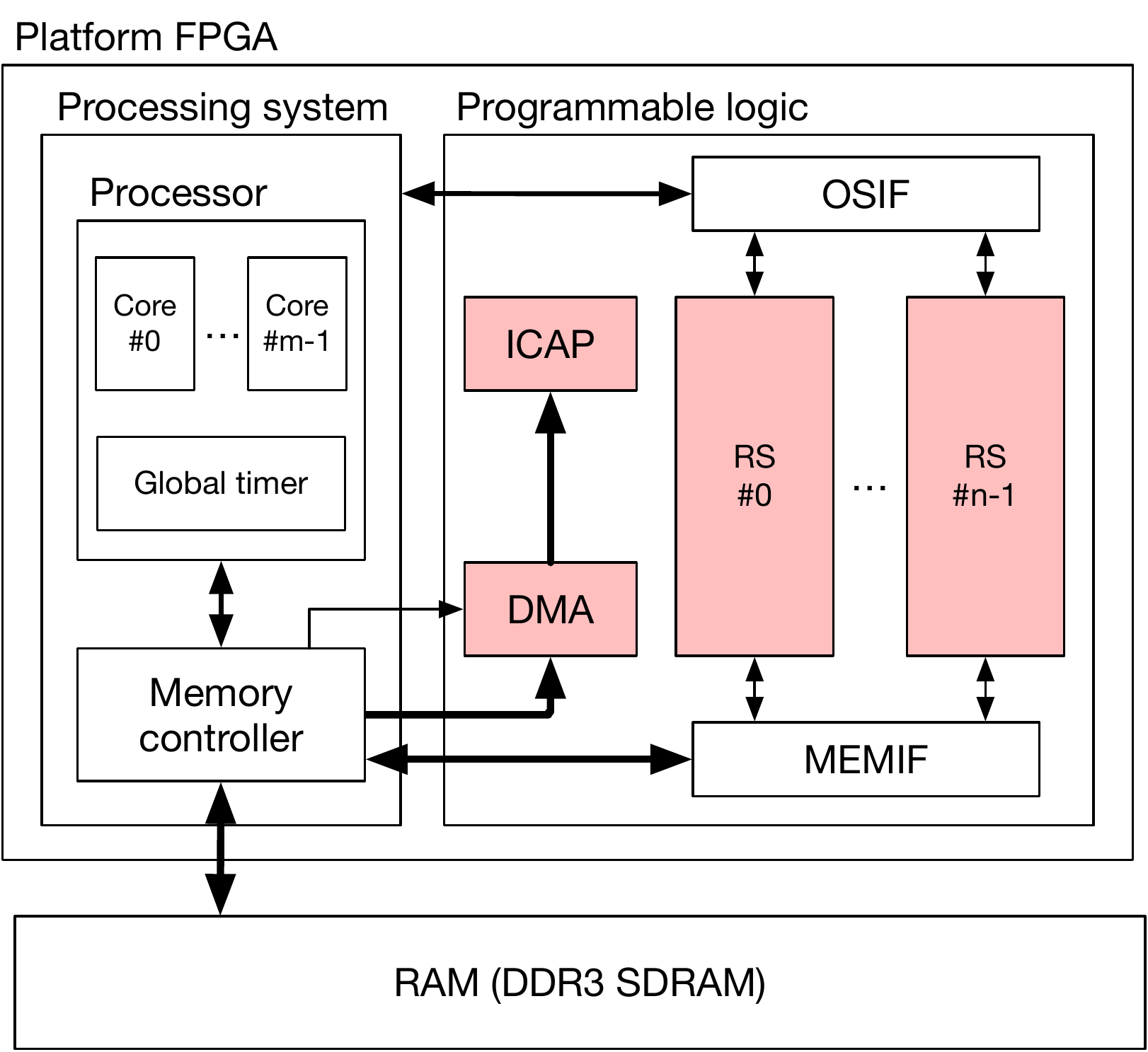}    
    \caption{Hardware architecture for the ReconROS executor}
    \label{fig:reconrosexecutorhwoverview}
\end{figure}

Figure~\ref{fig:reconrosexecutorhwoverview} highlights the architecture for ReconROS using our novel ROS 2 executor. According to the underlying ReconOS architecture, the programmable logic part of a platform FPGA contains a set of $n$ reconfigurable slots ($\mathrm{RS}$) that can accommodate hardware threads, which implement the hardware-mapped ROS 2 callbacks. The number and sizes of these reconfigurable slots is application-specific and configured during the design process. Each such reconfigurable slot is connected to an operating system interface (OSIF) for communication with the host operating system Linux running on the processor cores, and to a memory interface (MEMIF) for accessing shared external memory. Another component omitted Figure~\ref{fig:reconrosexecutorhwoverview} for simplicity is the memory subsystem that provides arbitration between the MEMIFs and includes a memory management unit to allow the hardware threads to work with virtual addresses.

ROS 2 hardware-mapped callbacks are loaded into reconfigurable slots on demand during runtime with a partial reconfiguration process utilizing the ZyCAP implementation~\cite{vipin2014zycap}. ZyCAP comprises an ICAP (internal configuration access port) interface and a DMA block on the hardware side, and a Linux kernel driver and user libraries on the software side. Library functions are available to load reconfigurable slots by setting up DMA transfers from external memory to the ICAP interface. 

We have chosen ZyCAP over using the processor configuration access port (PCAP) or the ICAP directly for two reasons: First, ZyCAP nominally features 3$\times$ higher performance for writing bitstreams, i.e., 382 MByte/s for the ZyCAP compared to 128 MBytes/s for PCAP~\cite{vipin2014zycap}. Second, ZyCAP includes a DMA controller that lowers CPU load for partial reconfiguration and, in turn, frees the CPU for executing ROS 2 callbacks. In the hardware architecture, the ZyCAP block is connected to the processing system (PS) via a high performance port (HPx) to transfer the bitstream and to an AXI-Lite interface for the configuration of DMA transactions.
From the existing ZyCAP project, we have simplified the user library functions to the three basic functions {\tt ZyCAP\_Init()} for the initialization, {\tt ZyCAP\_Write\_Bitstream()} for blocking write of the bitstream into the ICAP, and {\tt ZyCAP\_DeInit()} for de-initialization, and integrated them into the ReconROS library. Additionally, we have adapted the ZyCAP Linux kernel driver to more recent Linux kernels.

\subsection{Executor Design}

The main steps in designing the ReconROS executor are providing timers and coming up with a scheduling or dispatching algorithm, respectively, that utilizes all available processor cores and reconfigurable slots for callbacks. 

In the ROS 2 stack, timers are part of the high-level libraries rclcpp (C++) or rclpy (Python) and use the operating system to measure wall clock time. Since ReconROS builds on rcl, the underlying standard framework for ROS primitives, we have added a corresponding timer primitive to that. Our implementation uses the ARM Cortex-A9 global timer (see Figure~\ref{fig:reconrosexecutorhwoverview}) as its main time reference, and a low-overhead function {\tt ros\_timer\_is\_ready()} to check whether a time interval has expired.

The development of an executor algorithm is more challenging. In contrast to the standard ROS 2 executor (see Section~\ref{subsec:ROS2Executor}) that dispatches ready to execute callbacks to a number of identical software worker threads, typically one per available processor core, the ReconROS executor can either execute callbacks in software or hardware and, if executed in hardware, in specified reconfigurable slots. Therefore, our executor implementation is structured into an executor main thread, one software worker thread per processor core, and one hardware worker thread per reconfigurable slot. The main thread maintains four callback lists that include all callbacks registered at the executor, i.e., one for timers, one for subscribers, one for service servers, and the last one for service clients. Each entry in such a callback list comprises a unique identifier, a pointer to the received message in case of non-timer callbacks, and a {\it ResourceMask} that contains a field for software and each reconfigurable slot. If the execution mode is software, the corresponding field includes a function pointer to the callback code. If the execution mode is hardware, the corresponding fields contain pointers to callback bitstreams for the reconfigurable slots.

\begin{figure}[!h]
	\center
    \includegraphics[width=1\linewidth]{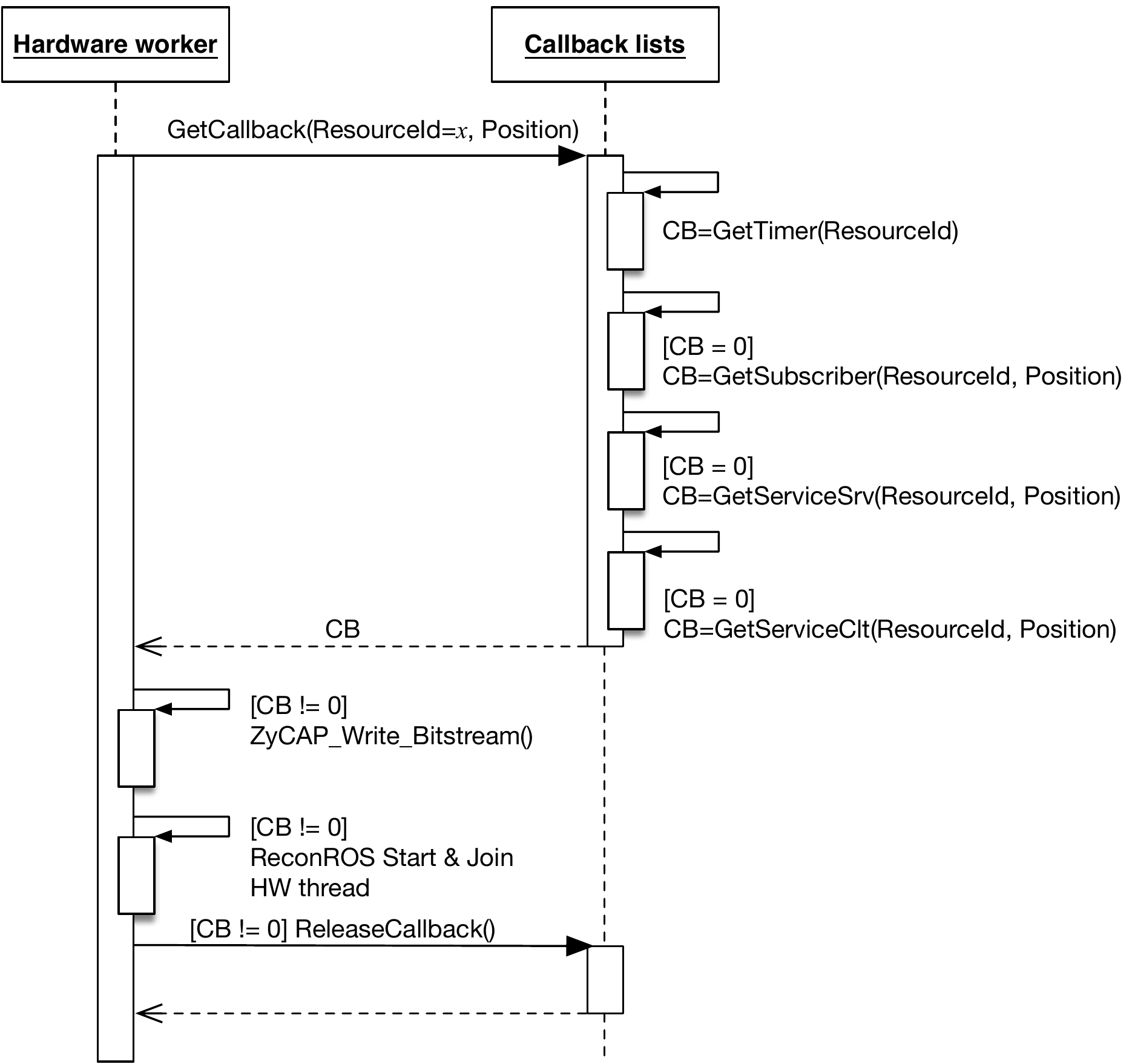}
    \caption{Sequence diagram for a hardware worker thread}
    \label{fig:hardwareworkersequencediagram}
\end{figure}

The overall $m$ software and $n$ hardware worker threads are started during the initialization of the executor. Each of these threads implements the inner loop of Figure~\ref{fig:ros2_standard_executor}. Figure~\ref{fig:hardwareworkersequencediagram} displays the functionality of the hardware worker thread for reconfigurable slot $x$. The thread accesses the callback lists in the order of timers, subscribers, service servers, and service clients and searches for ready callbacks (CB) with a matching entry $x$ in the {\it ResourceMask}. If such an entry is found (CB not zero), the thread checks wether the corresponding bitstream is already loaded in the reconfigurable slot $x$. If so, the callback is simply started; otherwise partial reconfiguration is performed to load the callback bitstream. The worker thread then waits until the callback is finished and runs into the next loop iteration. 

The standard ROS 2 executor shown in Figure~\ref{fig:ros2_standard_executor} collects ready non-timer callbacks before entering the inner loop. This ensures that all callbacks collected up to a certain time will actually be executed before new non-timer callbacks are considered. Very frequently appearing subscriber callbacks, for example, can thus not lead to starving callbacks for service servers and service clients. Since our ReconROS executor uses more independent worker threads, we resort to a different mechanism to avoid starvation. Each worker thread maintains an {\it OffsetVector} that holds for each non-timer callback list an entry {\it Position} that identifies the callback last served. Whenever the worker thread checks the list for the next ready callback, it starts the search from {\it Position$+1$}. After serving the callback, {\it Position} is incremented. {\it Position} is initialized with the length of the list and wrapped around to zero when the end of the callback list is reached. Software worker threads are identical, except that they start callback functions in software.

Our ReconROS executor tries to mimic the behavior of the ROS 2 standard executor and requires the designer to specify the size of the reconfigurable slots, and for each hardware callback the possible reconfigurable slots to which the callback can be mapped. Obviously, different and improved ReconROS executor designs are conceivable. For example, for callbacks that can be run in software and hardware, the executor could decide at runtime which mode to choose. Moreover, involved resource management problems arise if the reconfigurable slots are of different size and hardware callbacks are available for different reconfigurable slots. Such improved executor scenarios are left to future work.

\section{Design Example}
\label{sec:designexample}

As an example we elaborate on the ROS 2 application from Figure~\ref{fig:computational_graph} comprising six nodes. The ReconROS application from this design example handles the nodes {\tt /filter} and {\tt /inv\_kinematics}. The remaining nodes are assumed to be developed and compiled with ROS 2 design flows and mapped to other compute platforms. 

Both considered ROS 2 nodes comprise subscriber functionality for getting input data. According to the event-based programming approach, callbacks are invoked after the arrival of messages. In this example, the two callbacks are implemented in hardware and designed for  execution in a reconfigurable slot. 

Listing~\ref{lst:configuration_file} shows the ROS 2 related part of the {\sc ReconROS} configuration file for the overall ReconROS design project. The file starts with a block for the specification of the reconfigurable slot, which will accommodate the hardware callbacks. The specification is done by lists of contiguous resources for each type (LUT Slices, DSP, BRAM18, BRAM36). This lists can be derived by drawing pblocks using Xilinx Vivado and reading the resulting constraints.
The information for the ROS 2 nodes is organized into so-called resource groups. Lines 6--10 specify the {\tt /filter} node, beginning with the definition of a rosnode object named "/filter" in line 7. In line 8, the message object of type ROS 2 Image message is defined with further references to a ROS 2 message package ({\tt sensor\_msgs}) and the communication ({\tt msg}) as well as message types ({\tt Image}). Lines 9 and 10 declares primitives for the subscription of input data from topic {\tt /image\_raw} and the publication of filtered data to topic {\tt /image\_filtered}.

Lines 12--16 specify the {\tt /inv\_kinematics} node, including the rosnode object named "inv\_kinematics". Similar to the {\tt /filter} node, it comprises a message type definition (here inverse\_msg) and a publisher and subscriber for topic {\tt /ctrl\_out} and {\tt /tf\_out}.

\noindent\begin{minipage}{\linewidth}
    \centering
\begin{lstlisting}[floatplacement=H, numbers=left, numbersep=-5pt, frame=single, caption={Configuration file (Partial reconfiguration / ROS 2 related part) for the 
    {\sc ReconROS} application shown in Figure~\ref{fig:computational_graph}}\label{lst:configuration_file}]
   [HwSlot(at)ReconfSlot(0:0)]
   Id = 0
   Reconfigurable = true
   Region0 = SLICE_X0Y150SLICE_X103Y199, DSP48_X0Y60DSP48_X7Y79, RAMB18_X0Y60RAMB18_X5Y79, RAMB36_X0Y30RAMB36_X5Y39

   [ResourceGroup(at)ResourceGroupSobel]
   node_2 = rosnode, "/filter"
   img_msg = rosmsg, sensor_msgs, msg, Image 
   sub = rossub, node_2, img_msg, "/image_raw"
   pub = rospub, node_2, img_msg, "/image_filtered" 
 
   [ResourceGroup(at)ResourceGroupInverse]
   node_5 = rosnode, "/inv_kinematics"
   inverse_msg = rosmsg, std_msgs, msg, Uint32 
   sub = rossub, node_2, inverse_msg, "/ctrl_out"
   pub = rospub, node_2, inverse_msg, "/tf_out"     
\end{lstlisting}
\end{minipage}  

\noindent\begin{minipage}{\linewidth}
    \begin{lstlisting}[floatplacement=H,numbers=left, numberfirstline=true, numbersep=-5pt, frame=single,caption={C/C++ code (partial) for the HLS implementation of the subscriber callback for the {\tt /filter} ROS 2 node}\label{lst:hls_implementation_sobel}] 
    //Initdata contains pointer to message
    pMsg = THREAD_GETINITDATA();
    pMsg += OFFSETOF(sensor_msgs__msg__Image, 
        data.data);

    // Get pointer to image in memory and 
    // copy it to FPGA-internal BRAM
    MEM_READ(pMsg, pPayloadImage, 4);  
    MEM_READ(pPayloadImage[0], ram, 
        IMAGE_SIZE * 4); 

    SobelFilter(ram);

    // Write filtered image back to memory and 
    // publish filtered image
    MEM_WRITE(ram, pPayloadImage[0], 
        IMAGE_SIZE * 4); 
    ROS_PUBLISHER_PUBLISH(resourcesobel_pub, 
        resourcesobel_img_msg);

    THREAD_EXIT();
\end{lstlisting}
\end{minipage}

In Listing~\ref{lst:hls_implementation_sobel}, C/C++ code for the HLS implementation of the subscriber callback of the {\tt /filter} node is presented. The presented code is not performance optimized. %
The callback starts with accessing the initial data of the callback, which provides a pointer to the message object in main memory.
At this point, the message is already in the memory and ready for access. The position of the image data is calculated using the {\tt OFFSETOF} macro in line 3. Using the resulting pointer, the first use of {\tt MEM\_READ} macro reads the address of the image and then, with the second use of the macro, the callback reads the image into the {\tt ram} memory within in FPGA. After the execution of the Sobel filter function, the callback writes the filtered image back to main memory via the {\tt MEM\_WRITE} macro. After that, the callback publishes the filtered image by using the node-related publisher.

Listing~\ref{lst:hls_implementation_inverse} displays a similar procedure for the {\tt /inv\_kinematics} node, which basically relies on the same procedure as described in Listing~\ref{lst:hls_implementation_sobel}. Again, the received data from the subscriber is loaded into the FPGA-internal memory, processed, and then written back to main memory before publishing to the output topic.

\noindent\begin{minipage}{\linewidth}
\begin{lstlisting}[floatplacement=H,numbers=left, numbersep=-5pt, frame=single,caption={C/C++ code (partial) for the HLS implementation of the subscriber callback for the {\tt /inv\_kinematics} ROS 2 node}\label{lst:hls_implementation_inverse}]   
    //Initdata contains pointer to message
    pMsg = THREAD_GETINITDATA();

    // Get pointer to ctrl out in memory and 
    // copy it to FPGA-internal memory
    MEM_READ(pMsg, pPayloadInverse, 4);
    MEM_READ(pPayloadInverse[0], ram, 4);

    Inverse_Kinematics(ram);

    // Write outputdata to memory and publish result
    MEM_WRITE(ram, pPayloadInverse[0], 4);
    ROS_PUBLISHER_PUBLISH(resourcesinverse_pubdata, 
        resourcesinverse_inverse_msg);

    THREAD_EXIT();
\end{lstlisting}
\end{minipage}

The last needed user-created file for this example application is the main file, in which the ReconROS executor is instanciated and configured. The Listing~\ref{lst:implementation_main} shows the needed steps. In line 2, the ReconROS executor is initialized for execution without sw workers but with one hardware worker using the {\tt ReconROS\_Executor\_Init} function. The fourth argument for calling that function is the path to the partial bitstreams in the local filesystem. 
In line 3--4, the hardware callbacks are registered at the executor. The list of arguments comprises the executor instance, the ROS 2 node name, the {\it ResourceMask}, the ReconROS primitive type, the callback-creating ReconROS primitive instance and the ReconROS target message primitive. The last line of the code spins the executor and blocks until the application is terminated.

\noindent\begin{minipage}{\linewidth}
    \begin{lstlisting}[floatplacement=H,numbers=left, numbersep=-5pt, frame=single,caption={C/C++ code (partial) main thread of the ReconROS application}\label{lst:implementation_main}]   
    // Init the ReconROS executor without sw workers and one hw worker    
    ReconROS_Executor_Init(&reconros_executor, 0, 1, "/mnt/bitstreams/");

    ReconROS_Executor_Add_HW_Callback(&reconros_executor, "/inv_kinematics", 1, ReconROS_SUB, resourceinverse_subdata, resourceinverse_inverse_msg);
    ReconROS_Executor_Add_HW_Callback(&reconros_executor, "/filter", 1, ReconROS_SUB, resourcesobel_subdata, resourcesobel_image_msg);
    
    ReconROS_Executor_Spin(&reconros_executor);
    
\end{lstlisting}
\end{minipage}

\color{black}
\section{Experiments}
\label{sec:Exp}

This section reports on experiments to show the functionality of our ReconROS executor. First, we measure the execution times for a set of ROS 2 callbacks. Then, we determine the reconfiguration times for differently sized reconfigurable slots on our platform FPGA. Finally, we present a ROS 2 application comprising a desktop PC and a FPGA board and experiment with three different hardware/software mappings.

\subsection{Callback Execution Times}
\label{sec:Experiments:CBExecutionTimes}

As a first part of the evaluation of the ReconROS executor, we have measured the runtimes for five ROS 2 nodes, more precisely their callbacks:

\subsubsection*{Sobel filter} 
This callback implements a Sobel image filter~\cite{gonzalez2018digital} operating on three channels (RGB) of dimension $640 \times 480$. The filter applies two filter kernels on each channel of the image and calculates the absolute value of the dot product as an approximation for the geometric mean. The ROS 2 input and output messages are of the type {\tt Image} from the ROS 2 sensor message package.

\subsubsection*{Number sorting}
This callback provides a ROS 2 service which sorts an array of 32 Bit unsigned integers based on the odd-even transposition sort algorithm~\cite{knuth1998art}. The algorithm is based on a comparator network that employs $n$ stages with $n$ comparisons each to sort $n$ numbers. The ROS 2 node on the PC generates random numbers and publishes messages comprising 2048 numbers as an array. 

\subsubsection*{MNIST classifier}
This callback classifies handwritten digits from the MNIST dataset by implementing a neural network. The classifier is implemented using ROS 2 publish / subscribe communication. It subscribes for input images of size $28 \times 28$ and publishes the estimated digit as unsigned integer. The classifier consists of three convolution layers, three pooling layers and two fully connected layers. The achieved accuracy is about 97\%. 
\subsubsection*{Inverse kinematics}
This callback computes control signals for driving a servo motor that sets a joint angle based on a desired position and orientation of a robotic manipulation platform. The application is part of a larger mechatronic system for controlling the movements of a Stewart platform~\cite{stewartplatform} with six degrees of freedom. The computation involves coordinate transformations and an iterative implementation of the $\arctan()$ function. The ROS 2 input message is an unsigned 32 Bit integer packed with two fixed-point numbers in Q8.6 format that represent the desired rotation angles of the platform around the x-axis and the y-axis. The ROS 2 output messages is also a 32 Bit unsigned integer containing a 10 Bit unsigned integer which is the pulse width coded control signal for the motor.  

\subsubsection*{Hash calculation}
The hash calculation callback is implemented for the demonstration of a callback triggered by a periodic timer. At each run, the algorithm reads a $1920 \times 1080$ image with 24 bit color depth from main memory and calculates its SHA256 hash value. Afterwards, the hash value is published to a ROS 2 topic as an unsigned integer array with 8 elements.

\begin{table}[ht]
    \begin{center}
        \begin{tabular}{|l r r r|}
        \hline
        \makecell[l]{ROS 2 callback}      & \makecell[r]{$t_{exec-HW}$ \\ {[}ms{]}} & \makecell[r]{$t_{exec-SW}$ \\ {[}ms{]}} & Speedup \\[0.5ex] 
        \hline
        \makecell[l]{Sobel filter}            & 16.50      & 42.00       & 2.5\\
        \hline
        \makecell[l]{Number sorting}          & 0.85       & 41.00       & 48.2\\
        \hline
        \makecell[l]{MNIST classifier}        & 11.90      & 16.50       & 1.4\\
        \hline
        \makecell[l]{Inverse kinematics}      & 0.35       & 1.50         & 4.3\\
        \hline
        \makecell[l]{Hash calculation}        & 81.00      & 94.00       & 1.2\\
        \hline
        \end{tabular}
     \end{center}
    \caption{Execution times for five ROS 2 callbacks in hardware ($t_{exec-HW}$) and software ($t_{exec-SW}$), and the resulting speedup
	}
    \label{table:Experiments:callback_execution_times}
\end{table}

All callback functions have been coded in C/C++ and synthesized with Xilinx Vivado HLS to a Zynq Z7100 on a MiniITX FPGA board. The hardware callbacks run at 120 MHz, the ReconROS infrastructure at 100 MHz and the ARM Cortex-A9 at 666 MHz. Table~\ref{table:Experiments:callback_execution_times} shows the execution times for the callbacks, comprising the execution time for the callback function in software and the time between starting and completing the callback in hardware, respectively. The reconfiguration times are not included in this measurement. Speedups are achieved for all five callbacks, with the hash calculation resulting in the lowest and the number sorting resulting in the highest speedup.

\subsection{Reconfiguration Overheads}
\label{sec:Experiments:ReconfigOverheads}

For quantizing the reconfiguration time, we have created a ReconROS setup with four reconfigurable slots, $\mathrm{RS} \: \#0, \ldots, \mathrm{RS} \: \#3$.  
Table~\ref{table:Experiments:reconfiguration_time} shows the number of available resources per reconfigurable slot, the resulting bitstream size $S$, and the measured reconfiguration times $t_{rc}$ for the four reconfigurable slots.

\begin{table}[ht]
    \begin{center}
        \begin{tabular}{|l c c c r r|}
        \hline
        \makecell[c]{Reconfigurable \\ slot}   & \makecell[c]{Slice\\ LUTs} & DSPs & \makecell[c]{Block RAMs\\ (36 / 18)} & \makecell[c]{$S$ \\ {[}Byte{]}} & \makecell[c]{$t_{rc}$ \\ {[}ms{]}} \\[0.5ex] 
        \hline
        \makecell[c]{$\mathrm{RS}$ \#0} & 20800 & 160 & 60 / 120  & 2838976 & 24.0 \\
        \hline
        \makecell[c]{$\mathrm{RS}$ \#1} & 20800 & 160 & 60 / 120  & 2838976 & 24.0 \\
        \hline
        \makecell[c]{$\mathrm{RS}$ \#2} & 41600 & 320 & 240 / 120 & 5285728 & 38.4 \\
        \hline
        \makecell[c]{$\mathrm{RS}$ \#3} & 40800 & 280 & 200 / 100 & 4883328 & 36.9 \\
        \hline
        \end{tabular}
     \end{center}
    \caption{Reconfiguration slots with resources (Z7100 slices LUTs, DSPs, and Block RAMs), bitstream size and reconfiguration time}
    \label{table:Experiments:reconfiguration_time}
\end{table}

Using linear regression on the measured reconfiguration times and a reconfiguration time model that includes a constant offset part $t_{offset}$ and a bitstream size dependent part $S/B$, where $B$ denotes the transfer bandwidth, i.e., $t_{rc} = S / B + t_{offset}$, our measurements result in $t_{offset} = 6.8 ms$ and $B \approx 160 MByte/s$. The achieved bandwidth is much lower than the results reported in~\cite{vipin2014zycap}. The authors of~\cite{vipin2014zycap} apparently used a bare-metal implementation of the ZyCAP driver without operating system. Our current implementation suffers from copying the bitstream between user and kernel space. An improved implementation of the Linux driver with a zero-copy approach, e.g., based on {\it get\_user\_pages}, would increase the performance.

The reconfiguration times reported in Table~\ref{table:Experiments:reconfiguration_time} are directly depending on the size of the reconfigurable slot and the corresponding bitstream size, but they are basically independent on the hardware callback's functionality. The reconfiguration time adds to the execution time of a hardware callback only if the targeted reconfigurable slot is not yet configured with the required bitstream.

\subsection{Example Application}
\label{sec:Experiments:CBComparision}

In the last experiment we compare the performance of a ROS 2 application in software using the standard ROS 2 executor with two different hardware/software mappings using our ReconROS executor. The experimental setup comprises the Zynq Z7100 MiniITX FPGA board and a desktop PC with an Intel i7-8000 series CPU, connected via a Gigabit Ethernet connection. Both platforms run Ubuntu 18.04 LTS with ROS 2 dashing. All software components are compiled with optimization level O3. 

\begin{figure}[!h]
	\center
    \includegraphics[width=1\linewidth]{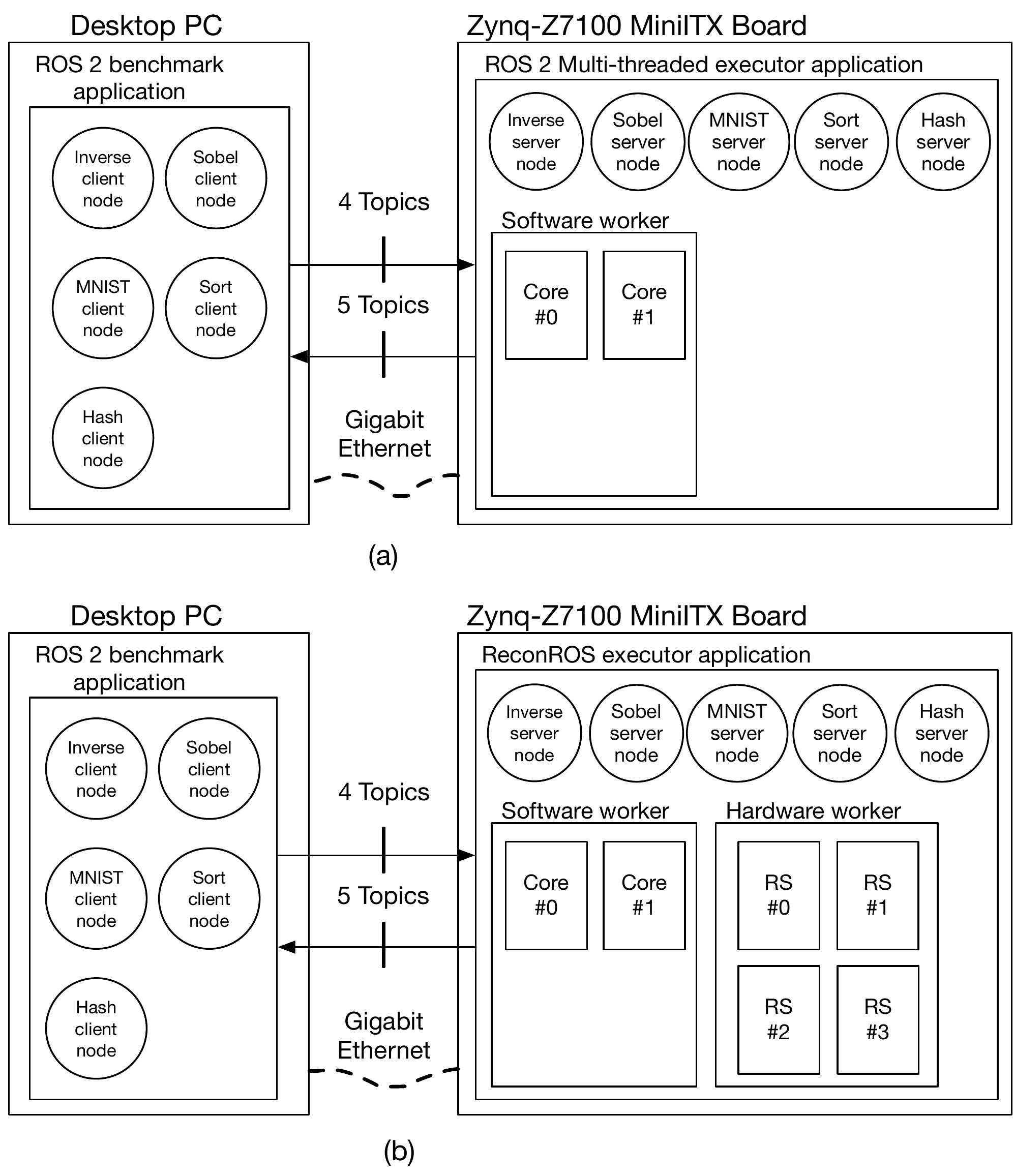}
    \caption{Experimental setup for a ROS 2 application with a standard ROS 2 executor (a), and our ReconROS executor (b)}
    \label{fig:evaluationsetup}
\end{figure}

\begin{figure*}[t]
	\centering
    \includegraphics[width=\textwidth]{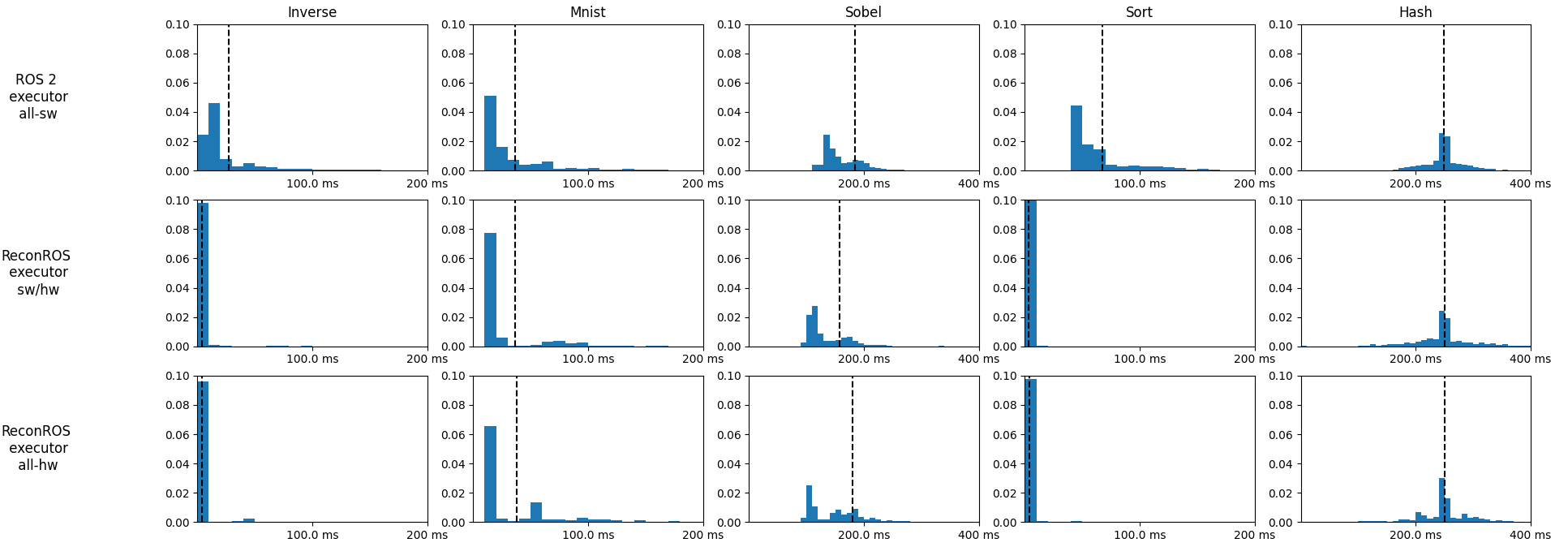}    
    \caption{Relative frequencies of the roundtrip times for the ROS 2 standard executor and two {\sc ReconROS} executor configurations; the dashed lines show the average roundtrip time for the specific ROS 2 node}
    \label{fig:executor_comparision}
\end{figure*}

The ROS 2 setup is illustrated in Figures~\ref{fig:evaluationsetup}(a) and (b). On the desktop PC, there are five ROS 2 client nodes programmed in C++ an compiled against the ROS 2 rclcpp library. These client nodes, i.e., the {\it Sort client}, {\it Inverse client}, {\it Sobel client}, and {\it MNIST client} nodes, comprise a publisher and a subscriber. Starting with an initial published message, the client's subscriber waits for a response from the corresponding server node on the FPGA. After receiving a new message for the topic, the clients immediately publish a new message for their server counterpart. The resulting roundtrip times are logged during the experiment. The {\it Hash client} node forms a special case. Since the hash server node only publishes messages, the client on the desktop PC just receives the messages and reports the times between consecutive messages.

Figure~\ref{fig:evaluationsetup}(a) sketches an all-software mapping, where a multi-threaded standard ROS 2 executor with two software worker threads on the FPGA dispatches the callbacks from the server nodes to two processor cores. Figure~\ref{fig:evaluationsetup}(b) displays the setup with the ReconROS executor and additional four hardware worker threads that dispatch callbacks to four reconfigurable slots. We have evaluated two mappings under the ReconROS executor, a mixed software/hardware mapping where the four callbacks with the highest speedups according to Section \ref{sec:Experiments:CBExecutionTimes}, i.e., {\it Number sorting}, {\it Inverse kinematics}, {\it Sobel filter}, and {\it MNIST classifier}, are executed in hardware and the {\it Hash calculation} callback is executed in software. The all-hardware mapping finally runs all callbacks in hardware.

Figure~\ref{fig:executor_comparision} analyzes the resulting roundtrip times for the three mappings. The figure plots for each of the five ROS 2 nodes and the three mappings the relative frequency over the roundtrip time. The dashed lines denote the averages. Going from the all-software over the software/hardware to the all-hardware mapping, the speedups based on the averaged roundtrip times are $6.21$ an $6.29$ for {\it Inverse kinematics}, $0.97$ and $1.00$ for the {\it MNIST classifier}, $1.03$ and $1.15$ for the {\it Sobel filter}, $18.18$ and $20.97$ for {\it Number sorting}, and $1.00$ for the {\it Hash calculation}. Overall, we make the following observations:

\begin{itemize}
\item The speedups for the individual ROS 2 nodes within the overall application follow the trends for the callbacks measured in isolation, shown in Table~\ref{table:Experiments:callback_execution_times}, although generally lower due to the communication between desktop PC and FPGA board, the ROS 2 communication layers, and the executors. For {\it Number sorting}, {\it Inverse kinematics}, and to some extent the {\it Sobel filter}, distinct speedups are realized. 

\item The hash calculation is triggered with a $250$ ms period. The distribution of roundtrip times shows entries with less and more than $250$ ms since the ROS 2 client on the desktop PC measures times between arriving messages from this callback. Some messages are delayed, which then likely reduces the time for the next message. 
\end{itemize}

\section{Conclusion and Future Work}
\label{sec:ConclusionFutureWork}
In this paper, we have introduced the ReconROS executor that enables event-based programming for hardware accelerated ROS 2 applications. In contrast to related work, the ReconROS executor leverages partial reconfiguration for loading hardware-mapped callbacks on demand to predefined reconfigurable slots on the FPGA's logic resource. Additionally, the ReconROS executor extends the ROS 2 standard executor and allows it to schedule or dispatch callbacks to software and hardware. 

In future work, we plan to expand the mapping and scheduling strategies of the ReconROS executor to optimize the hardware/software mapping and the resource management. In particular, callbacks that can run in both software and hardware will allow for taking runtime mapping decisions. With respect to resource management, techniques that minimize the unused resources within loaded slots or preloading of hardware callbacks are worth investigating.

\balance
\bibliographystyle{IEEEtran}
\bibliography{lienen22_fccm}

\end{document}